\documentclass{article}

\usepackage[final, nonatbib]{neurips_2022}

\usepackage[utf8]{inputenc} % allow utf-8 input
\usepackage[T1]{fontenc}    % use 8-bit T1 fonts
\usepackage{hyperref}       % hyperlinks
\usepackage{url}            % simple URL typesetting
\usepackage{booktabs}       % professional-quality tables
\usepackage{amsfonts}       % blackboard math symbols
\usepackage{amsmath}
\usepackage{amsthm}
\usepackage{CJKutf8}
\usepackage{graphicx}
\usepackage{nicefrac}       % compact symbols for 1/2, etc.
\usepackage{microtype}      % microtypography
\usepackage{xcolor}         % colors
\usepackage[style=numeric]{biblatex}
\usepackage[]{algorithm2e}
\usepackage{algorithmic}
\newtheorem{proposition}{Proposition}
\newtheorem*{proposition*}{Proposition}

\title{Equivariant Networks for Zero-Shot Coordination}

\author{%
  Darius Muglich \\
  University of Oxford\\
  \texttt{dariusm1997@yahoo.com} \\
  \And
  Christian Schroeder de Witt \\
  FLAIR, University of Oxford \\
  \texttt{cs@robots.ox.ac.uk} \\
  \And
  Elise van der Pol \\
  Microsoft Research AI4Science \\
  \texttt{evanderpol@microsoft.com} \\
  \And
  Shimon Whiteson \\
  University of Oxford \\
  \texttt{shimon.whiteson@cs.ox.ac.uk} \\
  \And
  Jakob Foerster \\
  FLAIR, University of Oxford \\
  \texttt{jakob.foerster@eng.ox.ac.uk} \\
}

\begin{document}

\maketitle

\begin{abstract}
 Successful coordination in Dec-POMDPs requires agents to adopt robust strategies and interpretable styles of play for their partner. A common failure mode is \textit{symmetry breaking}, when agents arbitrarily converge on one out of many equivalent but mutually incompatible policies. Commonly these examples include partial observability, e.g. waving your right hand vs. left hand to convey a covert message. In this paper, we present a novel equivariant network architecture for use in Dec-POMDPs that effectively leverages environmental symmetry for improving zero-shot coordination, doing so more effectively than prior methods. Our method also acts as a ``coordination-improvement operator'' for generic, pre-trained policies, and thus may be applied at test-time in conjunction with any self-play algorithm. We provide theoretical guarantees of our work and test on the AI benchmark task of Hanabi, where we demonstrate our methods outperforming other symmetry-aware baselines in zero-shot coordination, as well as able to improve the coordination ability of a variety of pre-trained policies. In particular, we show our method can be used to improve on the state of the art for zero-shot coordination on the Hanabi benchmark.
\end{abstract}

\section{Introduction}\label{sec:intro}

A popular method for learning strategies in partially-observable cooperative Markov games is via \textit{self-play} (SP), where a joint policy controls all players during training and at test time \cite{hu2019simplified, samuel1959some, tesauro1994td, yu2021surprising}. While SP can yield highly effective strategies, these strategies often fail in \textit{zero-shot coordination} (ZSC), where independently trained strategies are paired together for one step at test time. A common cause for this failure is mutually incompatible symmetry breaking (e.g. signalling $0$ to refer to a ``cat'' vs. a ``dog''). 
Specifically, this symmetry breaking results in policies that act \textit{differently} in situations which are equivalent with respect to the symmetries of the environment~\cite{hu2020other, shih2021critical}. 
% Here agents misinterpret one another merely due to differing choices in how each interacts with environmental symmetry \cite{hu2020other, shih2021critical}.
%
\begin{figure}
    \centering
    \includegraphics[width=100mm]{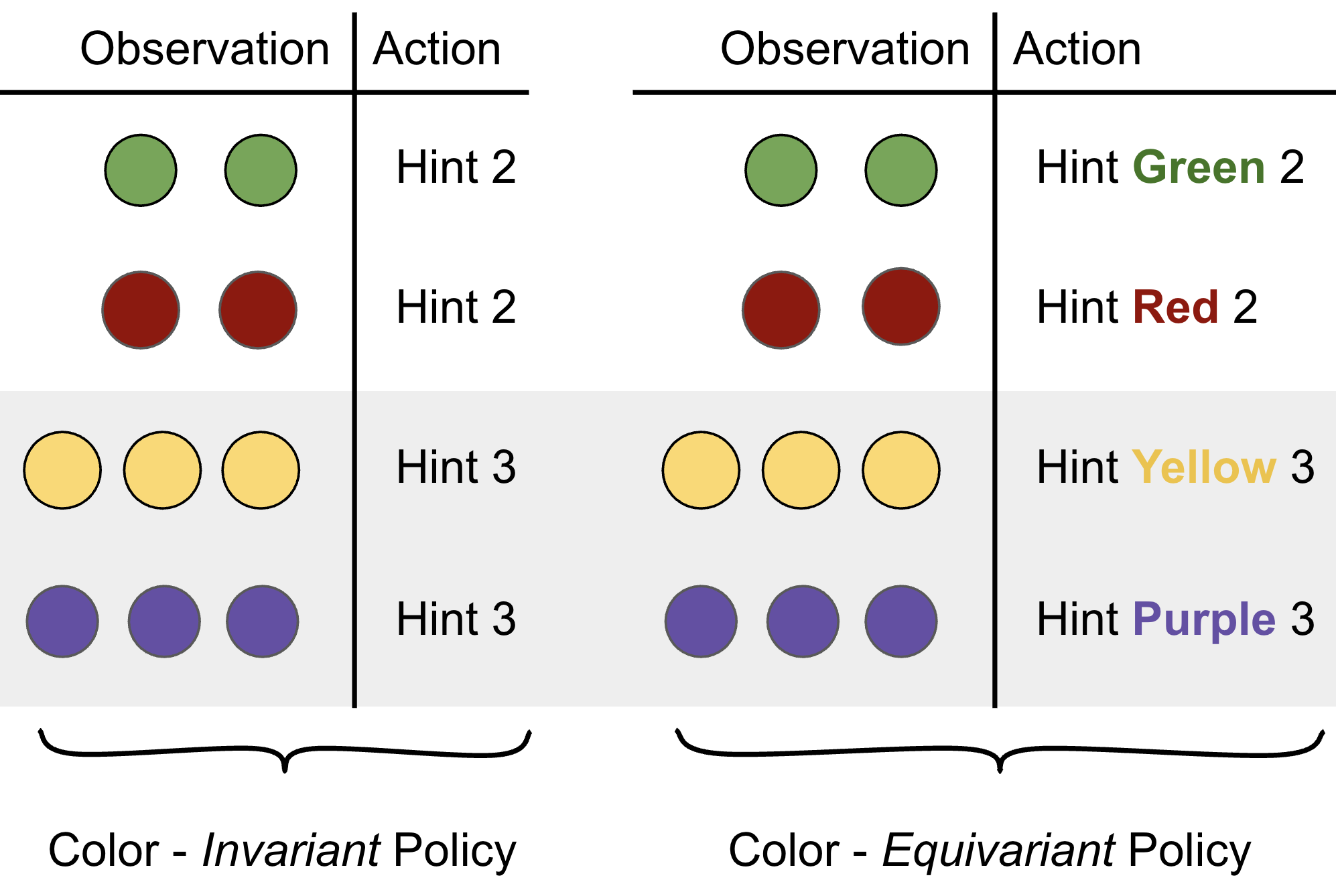}
    \caption{Illustrating different kinds of symmetry-robust policies, where the symmetries in this example are colors. Policies make actions based on the number of circles and the color. Invariant policies act irrespective of the change in color (i.e. only the number of circles matters), while equivariant policies act in correspondence with the change in color (i.e. changing the color will cause a corresponding change to the action).}
    \label{fig:what-is-equiv}
\end{figure}

To address this, we are interested in devising \textit{equivariant} strategies, which we illustrate in Figure \ref{fig:what-is-equiv}. Equivariant policies are such that symmetric changes to their observation cause a corresponding change to their output. In doing so, we fundamentally prevent the agent from breaking symmetries over the course of training.

In earlier work, the \textit{Other-Play} (OP) learning algorithm \cite{hu2020other} addressed symmetry in this setting by training agents to maximize expected return when randomly matched with symmetry-equivalent policies of their training time partners. This learning rule is implemented by independently applying a random symmetry permutation to the action-observation history of each agent during each episode. OP, however, has some pitfalls, namely: 1) it does not guarantee equivariance; 2) it poses the burden of learning environmental symmetry onto the agent; and 3) it is only applicable during \textit{training time} and cannot be used as a policy-improvement operator.

To address all of this, our approach is to encode symmetry not in the data, but in the \textit{network itself}.

There are many approaches to building equivariant networks in the literature. Most work focuses on analytical derivations of equivariant layers \cite{cohen2016group, weiler20183d, worrall2019deep, worrall2017harmonic}, which can prove time-consuming to engineer for each problem instance and inhibits rapid prototyping. Algorithmic approaches to constructing equivariant networks are restricted to layer-wise constraints \cite{cesa2021program, finzi2021practical, gordon2019permutation, van2020mdp}, which require computational overhead to enforce. Furthermore, for complex problems such as ZSC in challenging benchmarks such as Hanabi \cite{bard2020hanabi}, these approaches are insufficient because 1) they are not scalable; and 2) they usually do not allow for recurrent layers, which are the standard method for encoding action-observation histories in partially observable settings.

In this paper, we propose the \textit{equivariant coordinator} (EQC), which uses our novel, scalable approach towards equivariant modelling in Dec-POMDPs. Specifically, our main contributions are:
\begin{enumerate}
    \item Our method, EQC, that mathematically guarantees symmetry-equivariance of multi-agent policies, and can be applied solely at test time as a coordination-improvement operator; and
    \item Showing that EQC outperforms prior symmetry-robust baselines on the AI benchmark Hanabi.
    %\item Using EQC to map any deep, non-linear neural architecture (incl. recurrent networks) to one that is equivariant and capable in coordination.
\end{enumerate}
\section{Background}\label{sec:background}

This section formalises the problem setting (\ref{sec:dec-pomdp}), provides a brief primer to some group-theoretic notions (\ref{sec:group-theory}), introduces equivariance (\ref{sec:equivariance}), provides background to prior multi-agent work on symmetry (\ref{sec:op-symm}), and offers a small, concrete example to help tie these notions together and facilitate intuition (\ref{sec:example}). 

\subsection{Dec-POMDPs}\label{sec:dec-pomdp}

We formalise the cooperative multi-agent setting as a decentralized partially-observable Markov decision process (Dec-POMDP) \cite{oliehoek2012decentralized} which is a 9-tuple $(\mathcal{S},$ $\mathcal{N},$ $\{\mathcal{A}^i\}_{i=1}^n,$ $\{\mathcal{O}^i\}_{i=1}^n,$ $\mathcal{T},$ $\mathcal{R},$ $\{\mathcal{U}^i\}_{i=1}^n,$ $T,$ $\gamma$), for finite sets $\mathcal{S}, \mathcal{N},$ $\{\mathcal{A}^i\}_{i=1}^n,$ $\{\mathcal{O}^i\}_{i=1}^n$, denoting the set of states, agents, actions and observations, respectively, where a superscript $i$ denotes the set pertaining to agent $i \in \mathcal{N}=\{1 , \dots, n\}$ (i.e., $\mathcal{A}^i$ and $\mathcal{O}^i$ are the action and observation sets for agent $i$, and $a^i \in \mathcal{A}^i$ and $o^i \in \mathcal{O}^i$ are a specific action and observation of agent $i$). We also write $\mathcal{A}=\times_i \mathcal{A}^i$ and $\mathcal{O}=\times_i \mathcal{O}^i$, the sets of joint actions and observations, respectively. $s_t \in \mathcal{S}$ is the state at time $t$ and $s_t=\{s^k_t\}_k$, where $s_t^k$ is state feature $k$ of $s_t$. %(state features refer to the finitely-many enumerated characteristics that together specify the environment state)
$a_t \in \mathcal{A}$ is the joint action of all agents taken at time $t$, which  changes the state according to the transition distribution $s_{t+1} \sim \mathcal{T}(s_{t+1} \ | \ s_t, a_t)$. The subsequent joint observation of the agents is $o_{t+1} \in \mathcal{O}$, distributed according to $o_{t+1} \sim \mathcal{U}(o_{t+1} \ | \ s_{t+1}, a_t)$, where $\mathcal{U}=\times_i \mathcal{U}^i$; observation features of $o_{t+1}^i \in \mathcal{O}^i$ are notated analogously to state features; that is, $o_{t+1}^i=\{o_{t+1}^{i,k}\}_k$. The reward $r_{t+1} \in \mathbb{R}$ is distributed according to $r_{t+1} \sim \mathcal{R}(r_{t+1} \ | \ s_{t+1}, a_t)$. $T$ is the horizon and $\gamma \in [0,1]$ is the discount factor.

Notating $\tau_t^i=(a_0^i,o_1^i,\dots,a_{t-1}^i,o_t^i)$ for the action-observation history of agent $i$, agent $i$ acts according to a policy $a_t^i \sim \pi^i(a_t^i \ | \ \tau_t^i)$. The agents seek to maximize the return, i.e., the expected discounted sum of rewards:
\begin{equation}
J_T := \mathbb{E}_{p(\tau_T)}[\sum_{t'\leq T} \gamma^{t'-1} r_{t'}],
\end{equation}
where $\tau_t = (s_0, a_0, o_1, r_1, \dots, a_{t-1}, o_t, r_t, s_t)$
is the trajectory until time $t$.

\subsection{Groups}\label{sec:group-theory}
A group $G$ is a set with an associative binary operation on $G$, such that, relative to the binary operation, the inverse of each element in the set is also in the set, and the set contains an identity element. For our purposes, each element $g \in G$ can be thought of as an automorphism over some space $X$ (i.e. an isomorphism from $X$ to $X$), and the binary operation is function composition. The cardinality of the group is referred to as its order. If a subset $K$ of $G$ is also a group under the binary operation of $G$ (a.k.a.\ a subgroup), we denote it $K < G$.

A group action\footnote{To disambiguate between actions of mathematical groups and actions of reinforcement learning agents, we always refer to the former as ``group actions''.} is a function $G \times X \rightarrow X$ satisfying $ex = x$ (where $e$ is the identity element) and $(g \cdot h)x = g \cdot (hx)$, for all $g,h \in G$. In particular, for a fixed $g$, we get a function $\alpha_g : X \rightarrow X$ given by $x \mapsto gx$. $g$ can be thought to be ``relabeling'' the elements of $X$, which aligns with our definition for Dec-POMDP symmetry introduced Section \ref{sec:op-symm}.

Groups have specific mathematical properties, such as associativity, containing the inverse of each element, and closure (closure is implicit of the binary operation being over $G$ itself) which are necessary for proving Propositions \ref{proposition:symmetric-property} and \ref{proposition:fixing-property} and ensuring equivariance.

\subsection{Equivariance}\label{sec:equivariance}
Let $G$ be a group. Consider the set of all neural networks $\mathbf{\Psi}$ of a given architecture whose first and ultimate layers are linear; that is, $\mathbf{\Psi} := \{ \psi \ | \ \psi = h(\hat{\psi}(f)) \}$, where $f,h$ are linear functions, and $\hat{\psi}$ is a non-linear function. We say a network $\psi \in \mathbf{\Psi}$ is equivariant (with respect to a group $G$, or $G$-equivariant) if
\begin{align}\label{eq:equivariance-neural}
    \mathbf{K}_g \psi(\mathbf{x}) = \psi(\mathbf{L}_g \mathbf{x}), \text{ for all $g \in G, \mathbf{x} \in \mathbb{R}^d$},
\end{align}
where $\mathbf{L}_g$ is the group action of $g \in G$ on inputs, $\mathbf{K}_g$ is similarly the group action on outputs, and $d$ is the dimension of our data. If $\psi$ satisfies Equation \ref{eq:equivariance-neural}, then we say $\psi \in \mathbf{\Psi}_\text{equiv}$. In the context of reinforcement learning, one may consider $\psi$ as the neural parameterization of a policy $\pi$\footnote{We use $\psi$ and $\pi$ for this reason more or less interchangeably.}, so a policy is $G$-equivariant if
\begin{align}\label{eq:equivariance-policy}
    \pi(\mathbf{K}_g(a) \ | \ \tau) = \pi(a \ | \ \mathbf{L}_g(\tau)), \text{ for all $g \in G$ and for all $\tau$}. 
\end{align}
If for any choice of $g \in G$ we have that $\mathbf{K}_g = \mathbf{I}$, the identity function/matrix, then we say $\pi$ (or $\psi$) is invariant to $g$.
\subsection{Other-Play and Dec-POMDP Symmetry}\label{sec:op-symm}
The OP learning rule \cite{hu2020other} of a joint policy, $\pi = (\pi^1, \pi^2)$, is formally the maximization of the following objective (which we state for the two-player case for ease of notation):
\begin{equation}\label{eq:other-play}
    \pi^* = \text{arg}\max_{\pi} \mathbb{E}_{\phi \sim \Phi} J(\pi^1, \phi(\pi^{2})),
\end{equation}
where $\Phi$ is the class of equivalence mappings (symmetries) for the given Dec-POMDP, so that each $\phi \in \Phi$ is an automorphism of $\mathcal{S}, \mathcal{A}, \mathcal{O}$ whose application leaves the Dec-POMDP unchanged up to relabeling ($\phi$ is shorthand for $\phi = \{ \phi_\mathcal{S}, \phi_\mathcal{A}, \phi_\mathcal{O}\}$). The expectation in Equation \ref{eq:other-play} is taken with respect to the uniform distribution on $\Phi$. 

Each $\phi \in \Phi$ acts on the action-observation history as
\begin{equation}\label{eq:permute-aoh}
    \phi(\tau_t^i) = (\phi(a_0^i), \phi(o_1^i), \dots, \phi(a_{t-1}^i), \phi(o_t^i)),
\end{equation}
and acts on policies as
\begin{equation}\label{eq:symm-equiv}
    \hat{\pi} = \phi(\pi) \iff \hat{\pi}(\phi(a) \ | \ \phi(\tau)) = \pi(a \ | \ \tau).
\end{equation}
Policies $\pi, \hat{\pi}$ in Equation \ref{eq:symm-equiv} are said to be symmetry-equivalent to one another with respect to $\phi$. To correspond with Equation \ref{eq:equivariance-policy}, we have $\mathbf{L}_\phi = \{\phi_\mathcal{A}, \phi_{\mathcal{O}}\}$ and $\mathbf{K}_\phi = \phi_\mathcal{A}$.

%However, there is no guarantee that $\pi^*$ in Equation \ref{eq:other-play}  satisfies Equation \ref{eq:equivariance-policy} (for $G=\Phi$): $\pi^*$ acting optimally under any permutation of the state-action space does not tell us \textit{how} $\pi^*$ acts relative to each permutation (as Equations \ref{eq:equivariance-neural} and \ref{eq:equivariance-policy} do). Hence, different criteria for symmetry-robustness are possible (as we show in Section \ref{sec:results}, policies that satisfy Equation \ref{eq:equivariance-policy} and policies trained with respect to Equation \ref{eq:other-play} may be markedly different). This raises the question as to which criterion is best suited for a given application domain, and in particular, for our domain of cooperative multi-agent reinforcement learning and ZSC.

\subsection{Idealized Example}\label{sec:example}

Consider a game where you coordinate with an unknown stranger. Each of you must pick a lever from a set of $10$ levers. Nine of the levers have $1$ written on them, and one has $0.9$ written. If you and the stranger pick the same lever then you are paid out the amount that is written on the lever, otherwise you are paid out nothing.

We have $G:=S_9 \leq \Phi$, where $S_9$ is the symmetric group over $9$ elements (all possible ways to permute $9$ elements), since the nine levers with $1$ written on them are symmetric. If you (using policy $\pi^1$) pick the $1$-point lever $l_1$ and the stranger (using policy $\pi^2$) picks the $1$-point lever $l_2$, then $\pi^1$ and $\pi^2$ would be symmetry-equivalent with respect to any $\phi \in G$ that permutes the lever choices (e.g. $\phi(l_1) = l_2$, or equivalently $\mathbf{L}_\phi(l_1) = \mathbf{K}_\phi(l_1) = l_2$, as the levers are both observations and actions). If, as is likely, $l_1 \neq l_2$, you and the stranger leave empty-handed, illustrating symmetry breaking. If during training you and the stranger constrained yourselves to the space of equivariant policies whilst trying to maximize payoff, you would both converge to the policy that picks the $0.9$-point lever (which is not only equivariant, but invariant to all $\phi \in G$), and so the choice of using equivariant policies allows symmetry to be maintained and payout to be guaranteed.

\section{Method}\label{sec:method}

We now present our methodology and architectural choices, as well as several theoretical results.

\subsection{Architecture}\label{sec:architecture}
We first consider the \textit{symmetrizer} introduced in \cite{van2020mdp}, which is a functional that transforms linear maps to equivariant linear maps. Inspired by this we present a generalisation of the symmetrizer that maps from $\mathbf{\Psi}$ to the equivariant subspace,   $\mathbf{\Psi}_\text{equiv}$, namely $S : \mathbf{\Psi} \rightarrow \mathbf{\Psi}_\text{equiv}$, which we define as
\begin{align}\label{eq:symmetrizer-neural}
    S(\mathbf{\psi}) = \frac{1}{|G|} \sum_{g \in G} \mathbf{K}_g^{-1} \psi(\mathbf{L}_g), 
\end{align}
so that the first layer of $\psi$, denoted $f$, being linear, is composed with each $\mathbf{L}_g$, and the ultimate layer, denoted $h$, is similarly composed with $\mathbf{K}_g^{-1}$  (i.e. $\mathbf{K}_g^{-1}\psi(\mathbf{L}_g) = \mathbf{K}_g^{-1}\circ h(\hat{\psi}(f \circ \mathbf{L}_g))$, where $\psi = h(\hat{\psi}(f)$).

Below we establish properties of $S$ that were analogously shown for the linear symmetrizer case \cite{van2020mdp}:
\begin{proposition}\label{proposition:symmetric-property}
(Symmetric Property) $S(\psi) \in \mathbf{\Psi}_\text{equiv},$ for all $\psi \in \mathbf{\Psi}$; that is, $S$ maps neural networks to equivariant neural networks.
\end{proposition}
\begin{proposition}\label{proposition:fixing-property}
(Fixing Property) $\mathbf{\Psi}_\text{equiv} = \text{Ran}(S);$ that is, the range of $S$ covers the entire equivariant subspace.
\end{proposition}
The proofs of these propositions can be found in the Appendix. The proof of Proposition \ref{proposition:symmetric-property} relies on properties of the group structure (such as closure of the group); it is thus critical for the symmetrizer to sum over permutations that together form a group structure, and not any random collection of permutations.

Propositions \ref{proposition:symmetric-property} and \ref{proposition:fixing-property} show in combination that the architecture of $S(\psi)$ is indeed equivariant, and that optimizing over the class $\text{Ran}(S)$ (that is, keeping the permutation matrices\footnote{Any group can be viewed as a group of permutations, or as a group of ``permutation matrices''; this is known as the permutation representation of a group \cite{cayley1854vii}. Also, the terminology ``permutation matrix'' alludes to the implementation of group actions as matrix products.} $\mathbf{L}_g$ and $\mathbf{K}_g^{-1}$ for all $g$ frozen while updating the other weights) is equivalent to optimizing over $\mathbf{\Psi}_\text{equiv}$. The sum over each $g \in G$ of $S(\psi)$ is highly parallelizable \cite{herlihy2012art}, which allows for maintaining tractability. In this way we algorithmically enforce equivariance, mitigating the need to design equivariant layers by hand, as is typical \cite{cohen2016group, cohen2016steerable, weiler2019general, weiler20183d, worrall2019deep, worrall2017harmonic}.

We note that for a network containing LSTM \cite{hochreiter1997long} layers, $S$ will still map the network to an equivariant one no matter the values used for the hidden and cell states; indeed, Propositions \ref{proposition:symmetric-property} and \ref{proposition:fixing-property} hold regardless of how the hidden and cell states are treated. However, how the hidden and cell states are treated matter for the overall performance of the network (equivariance is not a sufficient condition itself for success in ZSC, consider the uniformly random policy). We ergo propose the following: for each hidden state $h_{t,g}$ and cell state $c_{t,g}$ produced by the LSTM layer over input $(\mathbf{L}_g(\tau), h_{t-1}, c_{t-1})$, we compute $h_t = \frac{1}{|G|}\sum_{g \in G} h_{t,g}$ and $c_t = \frac{1}{|G|}\sum_{g \in G} c_{t,g}$, and proceed inductively. This is a non-obvious design choice, since each $h_{t,g}$ and $c_{t,g}$ is the output of a non-linear function; a more obvious choice would be to choose $h_t = h_{t,e}$ and $c_t = c_{t,e}$ where $e \in G$ is the identity element, but to ensure symmetry-aware play we experimented with this averaging scheme to ``symmetrically adjust'' the short and long-term memories. Since all the permutations over the input are computed in parallel with matrix multiplication, this design choice retains tractability. We test this choice empirically in Section \ref{sec:results}. See the Appendix for further experimentation surrounding this design choice.

\subsection{Algorithmic Approach}
Based on Propositions \ref{proposition:symmetric-property} and \ref{proposition:fixing-property}, a ``naive approach'' to obtaining a $G$-equivariant agent, for a given group $G$, might then be as follows: 1) initialize a random seed $\psi \in \mathbf{\Psi}$; 2) map $\psi$ to $S(\psi)$; 3) train $S(\psi)$ using any self-play algorithm, updating the weights while keeping the permutation matrices $\mathbf{L}_g$ and $\mathbf{K}_g^{-1}$ for all $g$ frozen; 4) $S(\psi)$ is ready for inference. This naive approach is sound, but time and memory complexity inhibit scaling to large choices of $G$. For this reason we present the following two options (that together summarize EQC):
\begin{enumerate}
    \item As a first option, we introduce ``$G$-OP'', where during training we sample random subsets of $G$ for each mini-batch (i.e. we select a random $g \in G$ for each $\mathbf{x}$ in the mini-batch, so that $g$ group-theoretically acts on $\mathbf{x}$, and update the weights of $\psi$) and at test time, we use all the permutations of $G$ by deploying $S(\psi)$.
    \item As a second option, we train under any self-play algorithm and at test time deploy $S(\psi)$.
\end{enumerate}
With Option 1 we train over the permutations of $G$ at training time. With Option 2 we train under standard SP and use $S(\cdot)$ to obtain $S(\psi)$ at test time, which can be viewed as a coordination-improvement operator. We empirically validate both options in Section \ref{sec:results}. From here on, we refer to $S(\psi)$ as ``symmetrizing $\psi$''.
%\\
%
%\begin{algorithm}[H]\label{algo:training}
% \KwData{$\tau \sim p(\tau)$}
%\While{Self-Play Training} {
% Obtain $\tau$ from environmental interaction\;
% \If{Using $G$-OP learning rule}{
%    Select $g \in G$ uniformly at random\;
%    Update the weights of $\mathbf{K}_g^{-1}\psi( \mathbf{L}_g)$ using the gradient computed from $\tau$, keeping the permutation matrices frozen\;
%    }
%  \Else {
%    Update the weights of $\psi$ using the gradient computed from $\tau$\;
%    }
% }
% $S(\psi)$ is ready for inference.\
% \caption{Training $G$-equivariant agents. \sw{Format this more nicely.}}
%\end{algorithm}
%

We expect compatibility between the $G$-OP trained agents and the symmetrizer at test time, since at training time, these agents were trained over these permutations. For Option 2, where we have a network $\phi$ not trained with permutations, it is worthwhile to inspect $S(\phi)$; while we know $S(\phi)$ is equivariant, we know little else regarding the conventions and style of play it would use. While symmetry is a critical component of ZSC, other factors (such as choosing robust actions that are likely to succeed under minimal assumptions \cite{agranov2012beliefs, costa2006cognition}) are also significant. The following points justify why $S$ maps any network to one that is not only equivariant, but overall stronger at ZSC: 1) the network averages over how $\psi$ would act under each $g \in G$, thus ensuring play that accounts for multiple symmetries; 2) the group closure property fixes the architecture under any permutation in the group (see the proof of Proposition \ref{proposition:symmetric-property}), forcing consistent play across different symmetries. We verify the empirical efficacy of Options 1 and 2 in Section \ref{sec:results}.

What remains is the choice of the group $G$ for our $G$-equivariant agents. One might consider the choice of $G=\Phi$, but this in general scales factorially with the size of the Dec-POMDP (e.g. the automorphism group over an environment with $m$ symmetric features grows asymptotically with the number of permutations over $m$ elements), and so quickly becomes infeasible. Hence, we treat $G$ as a hyperparameter, and in Section \ref{sec:results} we analyze how different choices of $G$ affect ZSC. In any case, picking $G$ as a subgroup of $\Phi$ may be motivated in and of itself, when enumerating all possible symmetries of an environment is intractable, or if an environment is obfuscated such that some symmetries cannot be known (as is the case in many large, realistic domains). In this way, our method can relax the assumption of requiring access to all environment symmetries by flexibly choosing equivariance with respect to a subgroup.

\section{Experiments}\label{sec:results}

The primary test bed for our methodology is the AI benchmark task Hanabi \cite{bard2020hanabi}. Hanabi is a unique and challenging card game that requires agents to formulate informative implicit conventions in order to be successful. Hanabi has served as the primary test bed for many algorithms designed for ZSC \cite{cui2021k, hu2020other, hu2021off, lupu2021trajectory} due to its representation as a formidable Dec-POMDP task (see the Appendix for an introduction to Hanabi).

To make sense of the ensuing results, recall that a perfect score in Hanabi is $25$, and that ``bombing out'' refers to expending all three fuse tokens and ending the game with a score of $0$. Agents thus must be cautious not to expend all fuse tokens.

In Hanabi, the class of environment symmetries are the permutations of the card colors, because assuming there is no supplemental information, a permutation of the colors of the cards leaves the game unchanged. We therefore have $\Phi = S_5$, the symmetric group over $5$ elements (all possible permutations over $5$ elements). $5$ here represents the number of different colors. $|S_5| = 5! = 120$.

In this section we analyze differences in play and the relationship to environmental symmetry between our equivariant agents and OP agents (\ref{sec:op-vs-equiv}) and evaluate various pre-trained agents before and after symmetrization (\ref{sec:policy-improve}). These subsections evaluate Options 1 and 2, respectively (see Section \ref{sec:method}).

For further details of the training setup and hyperparameters used in the following experiments, please refer to the Appendix.

\subsection{Symmetry-Robust Agents}\label{sec:op-vs-equiv}
Here we compare the $12$ OP agents from \cite{hu2020other} with our equivariant agents. Specifically, we conduct the following ablation study: we fix two subgroups of $S_5:$ $C_5$ and $D_{10}$, which are the cyclic group of order $5$ and the dihedral group of order $10$, respectively; the geometric and/or algebraic interpretations of these groups is not important, for us groups purely serve as structured collections of permutations \cite{cayley1854vii}. Next, we train $10$ agents with the $C_5$-OP learning rule (see Section \ref{sec:method}), then map these $10$ agents to equivariant ones using the $C_5$-symmetrizer. Similarly, we train $10$ agents with the $D_{10}$-OP learning rule, then map them with the $D_{10}$-symmetrizer. Finally, we compare the ZSC performance between these three pools of $10$ agents. This is an ablation, because we control for the order of the group by considering both $C_5$ and $D_{10}$ (in fact, $C_5$ is a subgroup of $D_{10}$) so to observe the effect of group order. Each agent uses Recurrent Replay Distributed Deep Q-Networks (R2D2) \cite{kapturowski2018recurrent}, and is trained using value decomposition networks (VDN) \cite{sunehag2017value} and the simplified action decoding (SAD) algorithm \cite{hu2019simplified}. The average self-play scores achieved by the OP agents is $23.93 \pm 0.02$, of the $C_5$-equivariant agents is $23.96 \pm 0.03$, and the $D_{10}$-equivariant agents is $23.95 \pm 0.03$.
\begin{table}
 \caption{Comparing OP agents with our equivariant agents (each $G$-equivariant agent is trained using $G$-OP then symmetrized at test time) for their respective cross-play (XP) abilities. Each agent is trained with a different seed. Each pair of agents was evaluated over $5000$ games, with the total averages compiled here. The error bars are the standard error of the mean.}
 \label{table:op-vs-equiv}
 \centering
 \begin{tabular}{llll}
    \toprule
    %\cmidrule(r){1-3}
    \textbf{XP Stats} & OP & $C_5$-Equivariant & $D_{10}$-Equivariant \\
    \midrule
    Average Scores & 15.32 $\pm$ 0.65 & 15.45 $\pm$ 0.47 & 16.08 $\pm$ 0.42 \\
    Average Bombout Rate & 30.8\% & 24.3\% & 19.7\% \\
    \bottomrule
 \end{tabular}
\end{table}

Table \ref{table:op-vs-equiv} shows the findings of our ablation analysis. We see that both equivariant pools outperform the OP agents, in spite of the OP agents being trained over far more permutations than any of our equivariant agents: $|C_5| = 5, |D_{10}| = 10, |S_5| = 120$ (the OP agents are trained on permutations from $S_5$). This is consistent with earlier work where equivariant networks outperform data augmentation approaches~\cite{van2021multi, wang2022so, weiler20183d, winkels20183d, worrall2017harmonic, zhu2022sample}. We conduct a Monte Carlo permutation test \cite{dwass1957modified, eden1933validity} to compare the $D_{10}$-equivariant agent with the OP agents, for which we bound the derived $p$-value in a 99\% binomial confidence interval.

We also observe that the $D_{10}$-equivariant agents outperform the $C_5$-equivariant agents, which we largely attribute to $C_5 < D_{10}$: if we have a group $G < \Phi$, and choose $\phi \in \Phi$ uniformly at random, then $P(\phi \in G) = \frac{|G|}{|\Phi|}$, and so for two groups $G, H$ with $|G| \leq |H|$, we have $P(\phi \in G) \leq P(\phi \in H)$.

\subsection{Equivariance as a Policy-Improvement Operator}\label{sec:policy-improve}

Here we consider the extent to which test-time symmetrizing improves ZSC performance across diverse policy types: we train $10$ IQL agents \cite{tan1993multi}, and use the $12$ OP agents and $12$ SAD agents from \cite{hu2020other} and the $5$ agents from \cite{hu2021off} (level $5$), and examine the effect of transforming each of these self-play trained agents with our equivariant architecture at test time. This selection of agent types is diverse in terms of both overall performance level as well as the style of play. Furthermore, we consider agents that are both trained for cross-play (OP and OBL) and trained for the self-play setting (SAD and IQL), which is an important dimension to account for in consideration of EQC's applicability. All agents use R2D2 \cite{kapturowski2018recurrent}, and the SAD and OP agents also use VDN \cite{sunehag2017value}. The average self-play scores achieved by the SAD agents is $23.97 \pm 0.04$, the IQL agents is $23.15 \pm 0.02$, the OP agents is $23.93 \pm 0.02$, and the OBL agents is $24.20 \pm 0.01$.
\begin{table}
 \caption{Cross-play (XP) scores and bombout rates of various diverse policy types symmetrized at test time. Each agent is trained with a different seed. Each pair of agents was evaluated over $5000$ games, with the total averages compiled here. The error bars are the standard error of the mean.}
 \label{table:policy-improver-xp}
 \centering
 \begin{tabular}{lllll}
    \toprule
    \textbf{XP Stats} & SAD & IQL & OP & OBL-L5 \\
    \midrule
    W/o symmetrizer & 2.52 $\pm$ 0.34 & 10.53 $\pm$ 0.78 & 15.32 $\pm$ 0.65 & 23.77 $\pm$ 0.06\\
    $C_5$-symmetrized & 3.61 $\pm$ 0.38 & 13.57 $\pm$ 0.66 & 16.07 $\pm$ 0.59 & 23.77 $\pm$ 0.05\\
    $D_{10}$-symmetrized & 3.61 $\pm$ 0.39 & 13.62 $\pm$ 0.65 &  16.48 $\pm$ 0.53 & 23.89 $\pm$ 0.04\\
    \bottomrule
 \end{tabular}
 \begin{tabular}{lllll}
    \toprule
    \textbf{Bombout Rate Stats} & SAD & IQL & OP & OBL-L5\\
    \midrule
    W/o symmetrizer & 87.2\% & 39.4\% & 30.8\% & 0.56\%\\
    $C_5$-symmetrized & 81.9\% & 33.3\% & 26.1\% & 0.33\%\\
    $D_{10}$-symmetrized & 82.0\% & 33.3\% & 24.6\% & 0.22\%\\
    \bottomrule
 \end{tabular}
\end{table}

Table \ref{table:policy-improver-xp} summarizes our findings over the policy types, where we consider symmetrization with both the cyclic group of order $5$, $C_5$, and the dihedral group of order $10$, $D_{10}$. We find that mapping any of these networks through EQC with either choice of group improves its cross-play ability with novel agents at test time. We conduct Monte Carlo permutation tests \cite{dwass1957modified, eden1933validity} to compare the $D_{10}$-symmetrized agents with their unsymmetrized counterparts for each policy type, for which we bound each derived $p$-value in a 99\% binomial confidence interval. The $p$-values for the respective one-tailed tests for each of the SAD, IQL, OP and OBL agents are found to be within $(0, 0.00028]$, $(0, 0.00028]$, $(0, 0.00028]$, and $[0.0825, 0.0973]$ respectively, making the differences significant at the level $\alpha = 0.01$ for the SAD, IQL, and OP agents, and at the level $\alpha = 0.1$ for the OBL agents. Therefore, the improvement in coordination found by symmetrizing agents is highly statistically significant.

In particular, we see that OP agents improve in ZSC when symmetrized. In the proof for Proposition \ref{proposition:fixing-property}, we showed that equivariant networks are fixed under the symmetrizer. Since these OP agents were trained under the same permutations they are being symmetrized with, yet still improve upon symmetrization, this suggests that the data augmentation of OP does not lead to full equivariance, and so encoding environmental symmetry at the model level is necessary to improve overall performance.

It is especially significant that EQC maps networks not trained with permutations (SAD, IQL, OBL) to ones that are overall better at ZSC; we justify the mechanism by which it accomplishes this in Section \ref{sec:method}. In particular, we demonstrate that EQC can improve on OBL, the state of the art for zero-shot coordination on the Hanabi benchmark at the time of writing.

\begin{figure}
\hspace*{-1cm} 
    \centering
    \includegraphics[width=120mm]{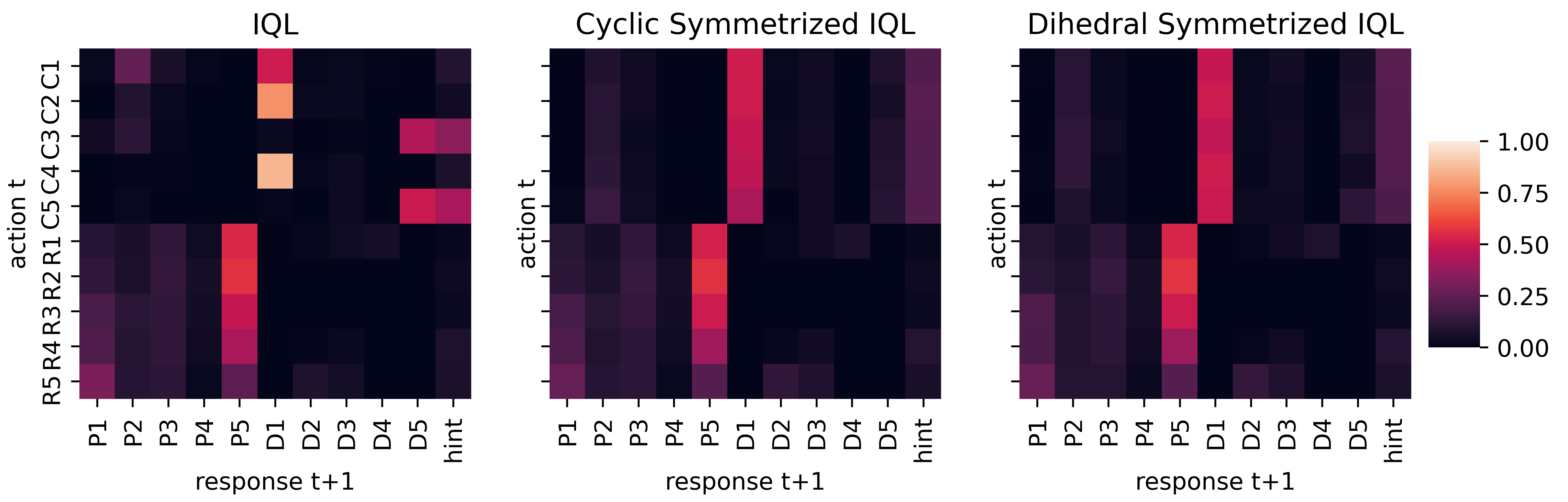}
    \caption{Conditional action matrices of IQL, i.e. $P(a_t^i \ | \ a_{t-1}^j)$, unsymmetrized (left) and symmetrized at test time (middle is $C_5$-symmetrized and right is $D_{10}$-symmetrized). The y-axis represents the action taken at timesetep $t$ and the x-axis shows the proportion of each action as response at timestep $t+1$. The matrices show the interactions between color/rank hinting and play/discarding. C1-5 and R1-5 mean hinting the 5 different colors and ranks respectively, and P1-5 and D1-5 mean playing and discarding the 1st-5th cards in the hand. We selected a random agent, and each plot is thereby computed by running 1000 episodes of self-play with the agent to compute the statistics.}
    \label{fig:cond_act_iql}
\end{figure}

Figure \ref{fig:cond_act_iql} shows the conditional action matrices (i.e. $P(a_t^i \ | \ a_{t-1}^j)$) of a symmetrized and unsymmetrized IQL agent (see the Appendix for the plots of other agents). This plot informs us on how differently the agent responds to possible actions of its partner, where large differences indicate specialized play, and therefore likely discrepancy between the conventions used by this agent and potential partners, which would be detrimental to cross-play. We see that the IQL agent has learned specialized conventions (for example, hinting the fourth color means for the partner to discard their first card), but that upon symmetrization its actions and responses become more consistent.

Another observation is that the $D_{10}$-symmetrizer does not always outperform the $C_5$-symmetrizer: namely for the case of the SAD agents in terms of bombout rate. The fact that $C_5 < D_{10}$ implies that the $D_{10}$-symmetrizer enforces equivariance across strictly more symmetires than the $C_5$ does, so the worse bombout rate may be attributed to the kind of play the symmetrized policy encourages, with a likely explanation as follows: if a policy is such that it acts very differently in each symmetry, then symmetrizing this policy over a large number of symmetry permutations will result in a policy with higher entropy (i.e. averaging over a large number of very different distributions results in one that is approximately uniform (see the Appendix)). But behaving ``randomly'' is not in general conducive to effective coordination, where one should try to use unambiguous conventions to facilitate play. \textit{Thus, when symmetrizing an arbitrary network, we tacitly assume some degree of regularity of the network across symmetries} (we control this regularity when we opt for the proposed $G$-OP learning rule). Nonetheless, both choices of $C_5$ and $D_{10}$ result in better ZSC than without any symmetrization for all the tested policy types (even with SAD, which learns notoriously specialized play \cite{hu2020other, hu2021off}), suggesting that policies in practice do generally satisfy these regularity requirements. This phenomenon, however, should be kept in mind and motivates future research into deriving the optimal choice of $G$ for $G$-equivariant agents.
%
%\begin{table}
%  \caption{OP vs Equivariant Training Regime under Different Group Permutations (removing the worst agent from each pool).}
%  \label{table:op-equiv-colorck}
%  \centering
%  \begin{tabular}{lll}
%    \toprule
%    %\cmidrule(r){1-3}
%    \textbf{$C_5$ Cross-Play Statistics}     & Cyclic OP     & Cyclic Equivariant \\
%    \midrule
%    Average Scores & 16.78 $\pm$ 0.04 & 13.45 $\pm$ 0.06 \\
%    %Non-Zero Scores & 16.36 $\pm$ 0.03 & 18.99 $\pm$ 0.01 \\
%    Proportion of cards played with known color & 12.9\% & 29.2\% \\
%    \bottomrule
%  \end{tabular}
%  \begin{tabular}{lll}
%    \toprule
%    %\cmidrule(r){1-3}
%    \textbf{$D_{10}$ Cross-Play Statistics}     & Dihedral OP     & Dihedral Equivariant \\
%    \midrule
%    Average Scores & 17.66 $\pm$ 0.03 & 18.01 $\pm$ 0.02 \\
%    %Non-Zero Scores & 19.22 $\pm$ 0.01 & 19.28 $\pm$ 0.01 \\
%    Proportion of cards played with known color & 14.0\% & 23.9\% \\
%    \bottomrule
%  \end{tabular}
%\end{table}
%
\section{Related Work}

\textbf{Zero-Shot Coordination.}
Zero-shot coordination is the problem of constructing independent agents that can effectively engage with one another in cross-play at test time. Many directions to this problem have been considered: \cite{cui2021k, hu2021off} explore controlling the cognitive-reasoning depth so to avoid formation of arbitrary conventions that can complicate cross-play. However, \cite{cui2021k} is computationally exacting, requiring days of GPU running time, and \cite{hu2021off} requires simulator (environment) access. \cite{shih2021critical} offers a modular approach to separately learn rule-dependent and convention-dependent behaviors to facilitate coordination, but their method comes at the expense of scalability, with experiments limited to bandit problems and 1-color Hanabi. \cite{parker2020ridge} uses repeated eigenvalues of the Hessian during training to find policies suited for ZSC, but is again inhibited by scalability, with experiments focused on lever games. \cite{nekoei2021continuous} investigate using life-long learning for ZSC, but their approach requires access to a pool of pre-trained policies. \cite{hu2020other}, which we have referred to at length throughout this paper, takes to data augmentation to train symmetry-robust agents. EQC can be used along with any of these approaches (even with \cite{hu2020other, hu2021off}, as we demonstrated in Section \ref{sec:results}), to encourage robust play.

\textbf{Ad-Hoc Teamwork.}
Similar to zero-shot coordination is the ad-hoc teamwork setting \cite{barrett2011empirical, muglich2022generalized, stone2010ad}, where an agent is pitted with an existing group of agents, and is tasked with learning to coordinate with this group during interaction. While ZSC and ad-hoc teamwork are closely related problems, they differ in that ad-hoc team play seeks policies that do well with \textit{arbitrary agents} (often an undefined problem setting), while  ZSC agents only need to perform well with agents that are optimized for ZSC using a common high-level approach.
In principle EQC can also be used to improve performance in ad-hoc teamplay, which we leave for future work.

\textbf{Equivariance.}
Equivariant networks are neural networks with symmetry constraints built in, such that for a transformation of the input, the output is appropriately transformed as well~\cite{anderson2019cormorant, cohen2016group, thomas2018tensor, weiler20183d, worrall2017harmonic}. Equivariant policies have shown improved data efficiency in single agent reinforcement learning~\cite{abdolhosseini2019learning, mondal2020group, park2021learning, van2020mdp, simm2020symmetry, wang2022so, zhu2021equivariant}. In the multi-agent case, prior works use permutation symmetries to ensure agent anonymity~\cite{bohmer2020deep, jiang2018graph, liu2020pic, sukhbaatar2016learning, sunehag2017value} or rotational equivariance over multiple agents~\cite{van2021multi}. While these approaches consider equivariance in multi-agent settings, they do not tackle the problem of ZSC. In terms of architecture, the symmetry ensemble of \cite{silver2016mastering} is related to ours, but we propose adaptations that tailor recurrent layers to multi-agent coordination, thus suiting our design not just for Markov Decision Processes (MDPs) \cite{bellman1957markovian} but for more general POMDPs \cite{astrom1965optimal} and Dec-POMDPs.

\section{Conclusion}\label{sec:conclusion}
In this work we proposed equivariant modelling for zero-shot coordination. To this aim we presented EQC, which we validated over the complex Dec-POMDP task of Hanabi. We showed that EQC greatly improves on \cite{hu2020other} by guaranteeing symmetry-equivariance of policies, and can be used as a policy-improvement operator to promote robust play for a diversity of agents. In this way, we also showed the extent to which symmetrizing improves overall performance in coordination. Furthermore, we showed that EQC is ``symmetry efficient'', such that it does not require access to all environment symmetries in order to perform well. In addition, we empirically validated that agents perform well under our proposed architectural choices.

There are many directions for future work. One is to consider how to derive optimal choices of the group $G$ for $G$-equivariant agents effective at ZSC, relative to a given policy type and task. Another direction is exploring how to efficiently uncover environment symmetries from a domain when they are not given nor assumed to be known. While our work focused on enforcing equivariance, future work could explore other fundamental aspects of coordination.

\section*{Acknowledgements}
The experiments were made possible by a generous equipment grant from NVIDIA. Christian Schroeder de Witt is generously funded by the Cooperative AI Foundation. We thank Jacob Beck for his constructive criticism of the manuscript.

\clearpage

\printbibliography

\clearpage

\appendix

\section{Equivariance Property Proofs}
Here we write the proofs for the Symmetric and Fixing Properties of our methodology introduced in Section \ref{sec:method}, which are analogous to those of \cite{van2020mdp}:
\begin{proposition*}
(Symmetric Property) $S(\psi) \in \mathbf{\Psi}_\text{equiv},$ for all $\psi \in \mathbf{\Psi}$; that is, $S$ maps neural networks to equivariant neural networks.
\begin{proof}
Without loss of generality take $g' \in G$ and $\psi \in \mathbf{\Psi}$. Then,
\begin{align*}
    \mathbf{K}_{g'}^{-1}S(\psi)(\mathbf{L}_{g'}\mathbf{x}) &= \mathbf{K}_{g'}^{-1}\frac{1}{|G|}\sum_{g \in G} \mathbf{K}_g^{-1} \psi(\mathbf{L}_g\mathbf{L}_{g'}\mathbf{x})
    \\&= \frac{1}{|G|}\sum_{g \in G} \mathbf{K}_{g'}^{-1}\mathbf{K}_g^{-1} \psi(\mathbf{L}_g\mathbf{L}_{g'}\mathbf{x})
    \\&= \frac{1}{|G|}\sum_{g \in G} \mathbf{K}_{gg'}^{-1} \psi(\mathbf{L}_{gg'}\mathbf{x})
    \\&= \frac{1}{|G|}\sum_{g'^{-1}m \in G} \mathbf{K}_{m}^{-1} \psi(\mathbf{L}_{m}\mathbf{x})
    \\&= \frac{1}{|G|}\sum_{m \in g'G} \mathbf{K}_{m}^{-1} \psi(\mathbf{L}_{m}\mathbf{x})
    \\&= \frac{1}{|G|}\sum_{m \in G} \mathbf{K}_{m}^{-1} \psi(\mathbf{L}_{m}\mathbf{x})
    \\&= S(\psi)(\textbf{x}).
\end{align*}
Thus $\mathbf{K}_{g'}S(\psi)(\mathbf{x}) = S(\psi)(\mathbf{L}_{g'}\mathbf{x})$.%Thus, $S(\psi)$ satisfies (\ref{eq:equivariance-neural}).
\end{proof}
\end{proposition*}
\begin{proposition*}
(Fixing Property) $\mathbf{\Psi}_\text{equiv} = \text{Ran}(S);$ that is, the range of $S$ covers the entire equivariant subspace.
\begin{proof}
By Proposition \ref{proposition:symmetric-property}, we have $\text{Ran}(S) \subset \mathbf{\Psi}_\text{equiv}$. It thus suffices to show $\mathbf{\Psi}_\text{equiv} \subset \text{Ran}(S)$. Take $\psi \in \mathbf{\Psi}_\text{equiv}$. Then,
\begin{align*}
    S(\psi) 
    &= \frac{1}{|G|} \sum_{g \in G} \mathbf{K}_g^{-1} \psi(\mathbf{L}_g)
    \\&= \frac{1}{|G|} \sum_{g \in G} \mathbf{K}_g^{-1} \mathbf{K}_g \psi
    \\&= \frac{1}{|G|} \sum_{g \in G} \psi
    \\&= \psi,
\end{align*}
where we have used Equation \ref{eq:equivariance-neural} since $\psi \in \mathbf{\Psi}_\text{equiv}$. Thus, $\psi$ is a fixed point of $S$, and so $\psi \in \text{Ran}(S)$.
\end{proof}
\end{proposition*}

\clearpage

\section{Optimal Meta-Equilibrium Proof}
\begin{proposition*}
Assume $G = \Phi$. If an agent chooses either the naive approach or the $G$-OP learning rule, then choosing either of these is payoff maximizing for the agent’s partner. In addition, both players choosing either the first approach or the $G$-OP learning rule is the best possible meta-equilibrium.
\begin{proof}
If an agent $\pi^1$ has an architecture $\pi^1 \in \text{Ran}(S)$ or uses the $G$-OP learning rule, then they necessarily have the objective to maximize
\begin{align*}
    \frac{1}{|G|} \sum_{\phi \in G} J(\pi^1, \phi(\pi^2)),
\end{align*}
because either of these choices forces consideration of all $\phi \in G$ equally. Since $G = \Phi$, we have
\begin{align*}
    \frac{1}{|\Phi|} \sum_{\phi \in \Phi} J(\pi^1, \phi(\pi^2))
    = \mathbb{E}_{\phi \in \Phi} J(\pi^1, \phi(\pi^2)),
\end{align*}
where the expectation is taken with respect to the uniform distribution of $\Phi$. We have thus derived the OP objective, and so what we sought so show is a corollary of \cite{hu2020other} (Proposition 2 in their paper).
\end{proof}
\end{proposition*}

\section{Averaging Over Very Different Distributions}
This passage concerns why if given a policy that acts very differently in different symmetries, which we then symmetrize with respect to a large group $G$, the resulting policy would be one that selects actions approximately uniformly at random.

Consider $l$ categorical distributions $\mu_1, \dots, \mu_l$ sampled uniformly from the collection of all categorical distributions over $k$ categories. Denoting $\mu_{i,j}$ as the probability of the $j^\text{th}$ category of distribution $i$, by symmetry in the categories and by assumption we have that $\mu_{a,r}$ and $\mu_{b,s}$ are equally distributed for all $1 \leq r,s \leq k$, $1 \leq a,b \leq l$. Therefore, by the law of large numbers, $\frac{1}{l} \sum_{i=1}^l \mu_{i,r} \approx  \frac{1}{l} \sum_{i=1}^l \mu_{i,s}$ when $l$ is large, for all $1 \leq r,s \leq k$. But since $\frac{1}{l}\sum_{i=1}^l \mu_i$ is a probability distribution, this implies $\frac{1}{l} \sum_{i=1}^l \mu_{i,j} \rightarrow \frac{1}{k}$ as $l$ becomes large, for all $1 \leq j \leq k$; i.e., $\frac{1}{l}\sum_{i=1}^l \mu_i$ approaches the uniform distribution over $k$ values.

The reader may identify $l = |G|$, $k$ as the number of legal actions, and each $\mu_i$ as the policy's action distribution under symmetry $i$, where we took each $\mu_i$ uniformly at random to mimic a policy that acts entirely differently per Dec-POMDP symmetry. While the uniform distribution over actions is indeed symmetry-equivariant (and invariant), it is not in general useful.

\section{Hanabi}
Hanabi is a cooperative card game that can be played with 2 to 5 people. Hanabi is a popular game, having been crowned the 2013 ``Spiel des Jahres'' award, a German industry award given to the best board game of the year. Hanabi has been proposed as an AI benchmark task to test models of cooperative play that act under partial information \cite{bard2020hanabi}. To date, Hanabi has one of the largest state spaces of all Dec-POMDP benchmarks.

The deck of cards in Hanabi is comprised of five colors (white, yellow, green, blue and red), and five ranks (1 through 5), where for each color there are three 1's, two each of 2's, 3's and 4's, and one 5, for a total deck size of fifty cards. Each player is dealt five cards (or four cards if there are 4 or 5 players). At the start, the players collectively have eight information tokens and three fuse tokens, the uses of which shall be explained presently.

In Hanabi, players can see all other players' hands but their own. The goal of the game is to play cards to collectively form five consecutively ordered stacks, one for each color, beginning with a card of rank 1 and ending with a card of rank 5. These stacks are referred to as fireworks, as playing the cards in order is meant to draw analogy to setting up a firework display\footnote{Hanabi (\begin{CJK*}{UTF8}{gbsn}花火\end{CJK*}) means `fireworks' in Japanese.}.

We call the player whose turn it is the active agent. The active agent must conduct one of three actions:

\begin{itemize}
    \item \textbf{Hint} - The active agent chooses another player to grant a hint to. A hint involves the active agent choosing a color or rank, and revealing to their chosen partner all cards in the partner's hand that satisfy the chosen color or rank. Performing a hint exhausts an information token. If the players have no information tokens, a hint may not be conducted and the active agent must either conduct a discard or a play.
    
    \item \textbf{Discard} - The active agent chooses one of the cards in their hand to discard. The identity of the discarded card is revealed to the active agent and becomes public information. Discarding a card replenishes an information token should the players have less than eight.
    
    \item \textbf{Play} - The active agent attempts to play one of the cards in their hand. The identity of the played card is revealed to the active agent and becomes public information. The active agent has played successfully if their played card is the next in the firework of its color to be played, and the played card is then added to the sequence. If a firework is completed, the players receive a new information token should they have less than eight. If the player is unsuccessful, the card is discarded, without replenishment of an information token, and the players lose a fuse token.
    
\end{itemize}

The game ends when all three fuse tokens are spent, when the players successfully complete all five fireworks, or when the last card in the deck is drawn and all players take one last turn. If the game finishes by depletion of all fuse tokens (i.e. by ``bombing out''), the players receive a score of 0. Otherwise, the score of the finished game is the sum of the highest card ranks in each firework, for a highest possible score of 25.

\section{Experiment Details}

Many of the hyperparameter choices are taken from \cite{hu2019simplified, hu2020other}. The main body of the network for each trained agent consists of $1$ fully connected layer with a hidden dimension of 512, 2 LSTM layers of 512 units, and two output heads for value and advantage, respectively. The networks are updated using the Adam optimizer \cite{kingma2014adam} with learning rate $6.25 \times 10^{-5}$ and $\epsilon = 1.5 \times 10^{-5}$. We use a batchsize of $128$ for training. The replay buffer size is $10^5$ episodes, and is warmed up with $10^4$ episodes before training commences. We use two GPUs for asynchronous actors to collect rollouts and record to an experience replay, and one GPU for computing gradients and model updates. The trainer sends its network weights to all actors every 10 updates and the target network is synchronized with the online network every $2.5 \times 10^3$ updates. Training was concluded after $10^3$ epochs, where each model reached (or near reached) convergence; each epoch consisted of $10^3$ updates. The compute we used were 3 NVIDIA GeForce RTX 2080 Ti GPUs and 40 CPU cores.

Our complete code for training and symmetrizing agents can be found in our GitHub repo: \url{https://github.com/gfppoy/equivariant-zsc}. This repo is based off the Off-Belief Learning codebase \cite{hu2021off} and the Hanabi Learning Environment (HLE).

\clearpage

\section{Comparing LSTM Modification choices}
Here we compare the two design choices mentioned in Section \ref{sec:architecture}: namely 1) $h_t = \frac{1}{|G|}\sum_{g \in G} h_{t,g}$ and $c_t = \frac{1}{|G|}\sum_{g \in G} c_{t,g}$ (\textit{averaging}); and 2) $h_t = h_{t,e}$ and $c_t = c_{t,e}$ (\textit{identity}). Table \ref{table:avg-vs-id} summarizes the results of symmetrizing the policies in Section \ref{sec:policy-improve} at test time using the dihedral group, using either the averaging or identity schemes. We can see the averaging scheme tends to slightly outperform the identity, an interesting finding given the counter-intuitive nature of the design choice.

\begin{table}[htb]\label{table:avg-vs-id}
 \caption{Cross-play (XP) scores and bombout rates of various diverse policy types symmetrized at test time. Agents are symmetrized with respect to the dihedral group, and we compare the averaging and identity schemes as described above. Each agent is trained with a different seed. Each pair of agents was evaluated over $5000$ games, with the total averages compiled here. The error bars are the standard error of the mean.}
 \label{table:avg-vs-id}
 \centering
 \begin{tabular}{lllll}
    \toprule
    \textbf{XP Stats} & SAD & IQL & OP & OBL-L5 \\
    \midrule
    W/o symmetrizer & 2.52 $\pm$ 0.34 & 10.53 $\pm$ 0.78 & 15.32 $\pm$ 0.65 & 23.77 $\pm$ 0.06\\
    $h_t = h_{t,e}$, $\text{ }c_t = c_{t,e}$ & 3.41 $\pm$ 0.41 & 13.32 $\pm$ 0.68 &  16.35 $\pm$ 0.54 & 23.88 $\pm$ 0.04\\
    $h_t = \frac{1}{|G|}\sum_{g \in G} h_{t,g}$, $\text{ }c_t = \frac{1}{|G|}\sum_{g \in G} c_{t,g}$ & 3.61 $\pm$ 0.39 & 13.62 $\pm$ 0.65 &  16.48 $\pm$ 0.53 & 23.89 $\pm$ 0.04\\
    \bottomrule
 \end{tabular}
\end{table}

\section{Conditional Action Matrices}

\begin{figure}[hbt!]
\hspace*{-1cm} 
    \centering
    \includegraphics[width=100mm]{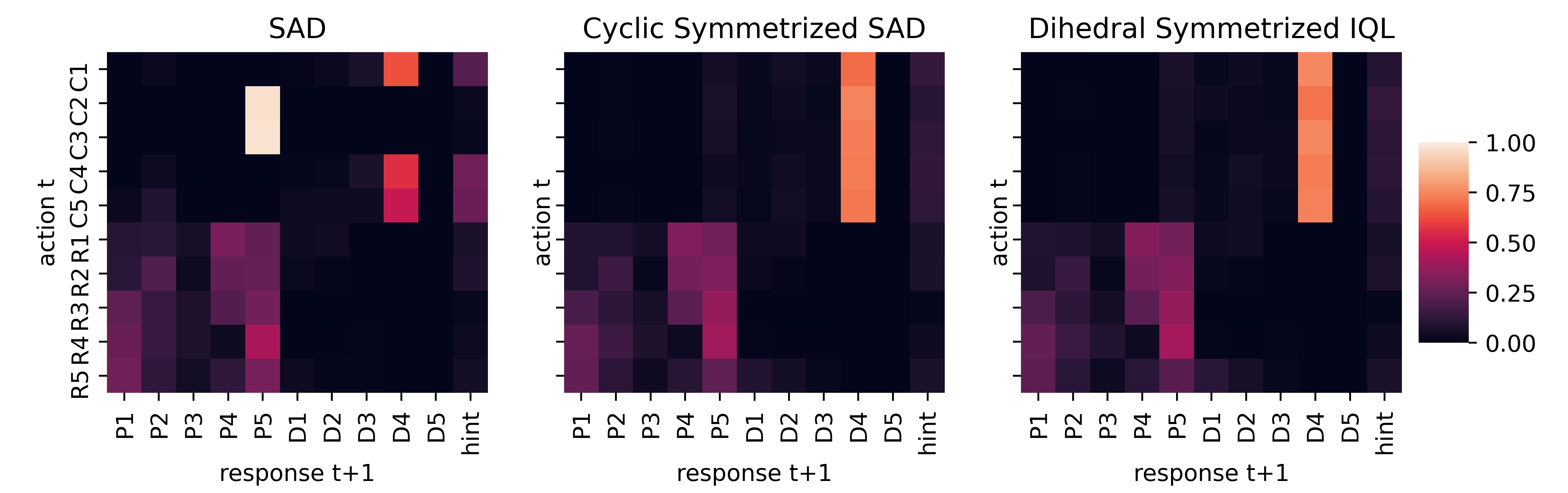}
    \caption{Conditional action matrices of SAD, i.e. $P(a_t^i \ | \ a_{t-1}^j)$, unsymmetrized (left) and symmetrized at test time (middle is $C_5$-symmetrized and right is $D_{10}$-symmetrized). The y-axis represents the action taken at timesetep $t$ and the x-axis shows the proportion of each action as response at timestep $t+1$. The matrices show the interactions between color/rank hinting and play/discarding. C1-5 and R1-5 mean hinting the 5 different colors and ranks respectively, and P1-5 and D1-5 mean playing and discarding the 1st-5th cards in the hand. We selected a random agent, and each plot is thereby computed by running 1000 episodes of self-play with the agent to compute the statistics.}
    \label{fig:cond_act_sad}
\end{figure}

\begin{figure}[hbt!]
\hspace*{-1cm} 
    \centering
    \includegraphics[width=100mm]{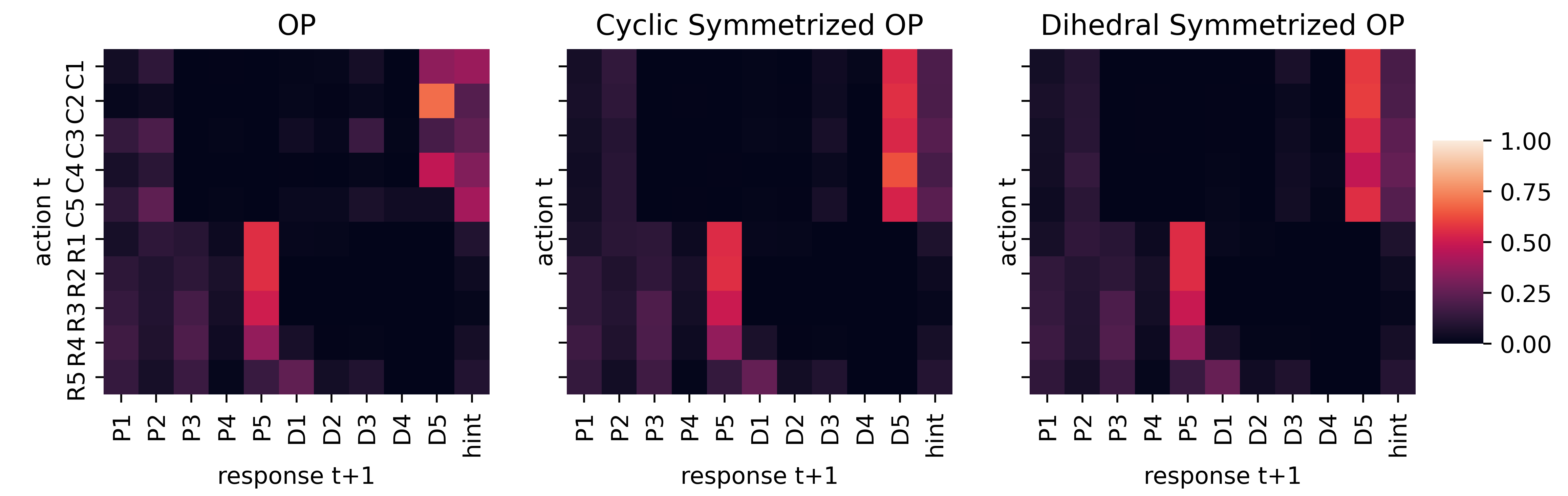}
    \caption{Conditional action matrices of OP, i.e. $P(a_t^i \ | \ a_{t-1}^j)$, unsymmetrized (left) and symmetrized at test time (middle is $C_5$-symmetrized and right is $D_{10}$-symmetrized). The y-axis represents the action taken at timesetep $t$ and the x-axis shows the proportion of each action as response at timestep $t+1$. The matrices show the interactions between color/rank hinting and play/discarding. C1-5 and R1-5 mean hinting the 5 different colors and ranks respectively, and P1-5 and D1-5 mean playing and discarding the 1st-5th cards in the hand. We selected a random agent, and each plot is thereby computed by running 1000 episodes of self-play with the agent to compute the statistics.}
    \label{fig:cond_act_op}
\end{figure}

\begin{figure}[hbt!]
\hspace*{-1cm} 
    \centering
    \includegraphics[width=70mm]{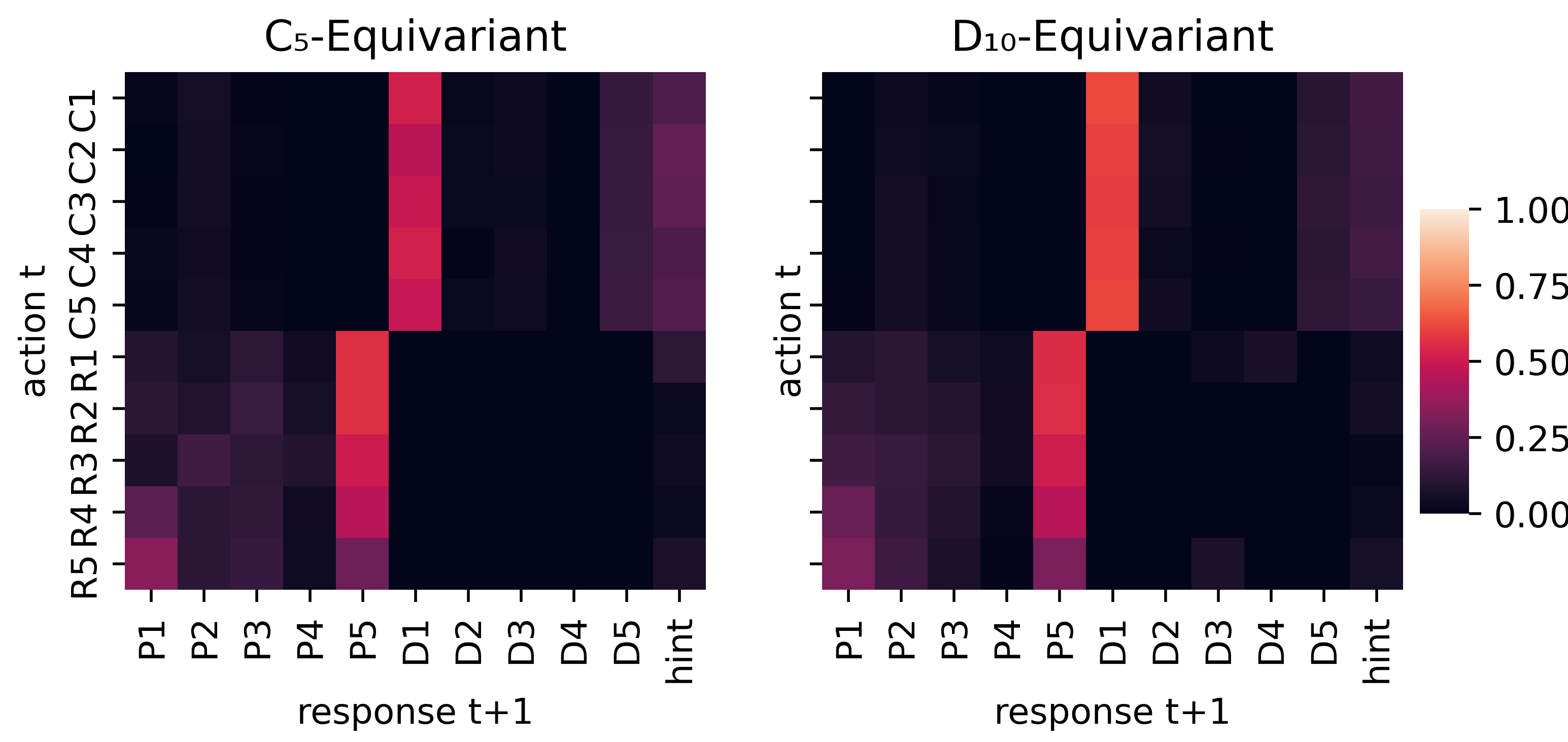}
    \caption{Conditional action matrices of $G$-equivariant agents, i.e. $P(a_t^i \ | \ a_{t-1}^j)$, $C_5$ on left and $D_{10}$ on right. The y-axis represents the action taken at timesetep $t$ and the x-axis shows the proportion of each action as response at timestep $t+1$. The matrices show the interactions between color/rank hinting and play/discarding. C1-5 and R1-5 mean hinting the 5 different colors and ranks respectively, and P1-5 and D1-5 mean playing and discarding the 1st-5th cards in the hand. We selected a random agent $C_5$-equivariant agent and a random $D_{10}$-equivariant agent, and each plot is thereby computed by running 1000 episodes of self-play with each agent to compute the statistics.}
    \label{fig:cond_act_equiv}
\end{figure}

\section{On the High Variance of Cross-Play Scores}
Reinforcement learning is notoriously sensitive to hyperparameter settings and seeds, making reproducibility of results challenging\footnote{https://media.neurips.cc/Conferences/NIPS2018/Slides/jpineau-NeurIPS-dec18-fb.pdf}. Correspondingly, in our experimentation we have found that average cross-play scores of groups of trained agents can vary greatly from group to group, where adjusting such hyperparameters as batch size, number of GPUs used for simulation, network architecture and seeds used can all lead to different average cross-play scores. This should be borne in mind for future research that aims to compare new results with existing baselines.

\section{Erratum}
In the original publication of our manuscript, a typographical error was made in the formal definition of symmetry-equivalent policies. Specifically, the described relationship between policies $\hat{\pi}$ and $\pi$, under a symmetry transformation $\phi$, was inaccurately represented with an inverse that should not have been applied. The correct relationship is:

$\hat{\pi} = \phi(\pi) \iff \hat{\pi}(\phi(a) \ | \ \phi(\tau)) = \pi(a \ | \ \tau)$

We reemphasize our primary methodological contribution focuses on the enforcement of equivariance as a means to enhance coordination in zero-shot settings. This approach underpins our results and the demonstration of state-of-the-art performance in zero-shot coordination tasks.

The corrections do not alter the core findings or the validity of the experiments and results presented. The demonstration of our method's effectiveness in improving coordination and achieving state-of-the-art performance in the domain of zero-shot coordination remains unchanged. We regret any confusion caused by these errors and appreciate the opportunity to clarify our contributions to the field.

\section{Broader Impact}
We have found that equivariant networks can effectively facilitate coordination and encourage play suited for cooperative settings. No technology is safe from being used for malicious purposes, which equally applies to our research. However, fully-cooperative settings target benevolent applications.

\end{document}